\documentclass{article}

\usepackage[preprint]{neurips_2025}

\usepackage{array}
\usepackage{graphicx}
 \usepackage{amsmath} 
\usepackage{multirow}
\usepackage{stfloats}
\usepackage{url}
\usepackage{verbatim}
\usepackage{graphicx}
\usepackage{graphicx}
\usepackage{amssymb}
\usepackage{booktabs}
\usepackage{xcolor}
\usepackage{colortbl}
\usepackage{hyperref}
\usepackage[capitalize]{cleveref}
\Crefname{section}{Sec.}{Sections}
\Crefname{table}{Tab.}{Tables}
\Crefname{figure}{Fig.}{Figures}
\definecolor{my_color}{HTML}{fbe0e0}

\usepackage{hyperref}       % hyperlinks

\usepackage[utf8]{inputenc} % allow utf-8 input
\usepackage[T1]{fontenc}    % use 8-bit T1 fonts
\usepackage{url}            % simple URL typesetting
\usepackage{booktabs}       % professional-quality tables
\usepackage{amsfonts}       % blackboard math symbols
\usepackage{nicefrac}       % compact symbols for 1/2, etc.
\usepackage{microtype}      % microtypography
\usepackage{xcolor}         % colors
\usepackage{wrapfig}

\newcommand{\ourmethod}{SIPL}
\newcommand{\promptir}{PromptIR}
\newcommand{\adair}{AdaIR}

\title{Boosting All-in-One Image Restoration via Self-Improved Privilege Learning
}

\author{%
Gang Wu \\
Harbin Institute of Technology\\
\texttt{gwu@hit.edu.cn} \\ 
\And
Junjun Jiang\thanks{Corresponding author.}\\
Harbin Institute of Technology\\
\texttt{jiangjunjun@hit.edu.cn} \\ 
\And
Kui Jiang\\
Harbin Institute of Technology\\
\texttt{jiangkui@hit.edu.cn} \\ 
\And
Xianming Liu\\
Harbin Institute of Technology\\
\texttt{csxm@hit.edu.cn} \\ 
}

\begin{document}

\maketitle

\begin{abstract}
Unified image restoration models for diverse and mixed degradations often suffer from unstable optimization dynamics and inter-task conflicts. This paper introduces Self-Improved Privilege Learning (SIPL), a novel paradigm that overcomes these limitations by innovatively extending the utility of privileged information (PI) beyond training into the inference stage. Unlike conventional Privilege Learning, where ground-truth-derived guidance is typically discarded after training, SIPL empowers the model to leverage its own preliminary outputs as \emph{pseudo-privileged signals} for iterative self-refinement at test time. Central to SIPL is Proxy Fusion, a lightweight module incorporating a learnable Privileged Dictionary. During training, this dictionary distills essential high-frequency and structural priors from privileged feature representations. Critically, at inference, the \emph{same} learned dictionary then interacts with features derived from the model's initial restoration, facilitating a self-correction loop. SIPL can be seamlessly integrated into various backbone architectures, offering substantial performance improvements with minimal computational overhead. Extensive experiments demonstrate that SIPL significantly advances the state-of-the-art on diverse all-in-one image restoration benchmarks. For instance, when integrated with the PromptIR model, SIPL achieves remarkable PSNR improvements of +4.58~dB on composite degradation tasks and +1.28~dB on diverse five-task benchmarks, underscoring its effectiveness and broad applicability. Codes are available at our project page \url{https://github.com/Aitical/SIPL/tree/main}.
\end{abstract}

\section{Introduction}\label{sec:introduction}
\vspace{-10pt}

Image restoration, a fundamental task within computer vision, seeks to reconstruct high-quality (HQ) images from degraded observations~\cite{Banham1997digitalsurvey, su2022survey}. Recently, the research community has increasingly focused on \emph{all-in-one} image restoration methods, designed to address multiple degradation types with a unified model~\cite{jiang2024survey}. Despite their significant practical value, consolidating diverse restoration capabilities into a single network introduces substantial optimization challenges. Consistent with previous studies~\cite{MioIR,wu2024harmony}, the all-in-one restoration task, typically a complex multi-task learning problem, is notably vulnerable to inter-task conflicts and negative transfer phenomena. Why does this occur? The core issue lies in the fundamental incompatibility between shared feature representations and task-specific requirements across different degradation types. For instance, the optimal feature representations for denoising (focusing on local texture patterns) may fundamentally differ from those required for dehazing (emphasizing global atmospheric light estimation). When forced to share network parameters, gradients originating from these distinct tasks often conflict, causing certain tasks, particularly those with larger gradient magnitudes or simpler optimization landscapes, to dominate the training process. This dynamic consequently results in imbalanced convergence rates and diminished overall performance.

\begin{figure}[t]
    \centering
    \includegraphics[width=\linewidth]{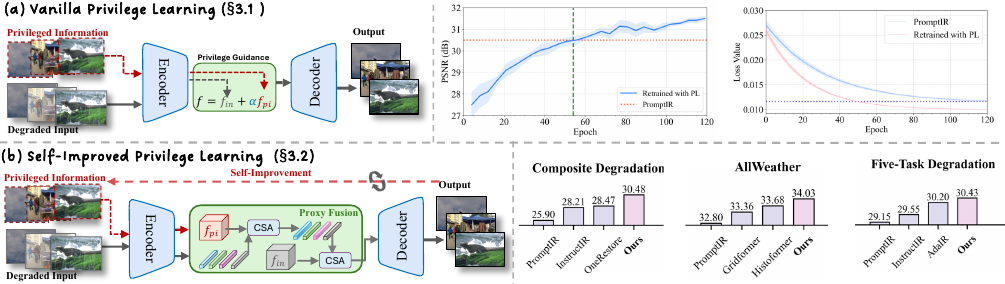}
    \vspace{-20pt}
    \caption{Conceptual comparison of learning frameworks: (a) Privilege Learning (PL) leverages privileged information during training for improved optimization. (b) Our proposed Self-Improved Privilege Learning (SIPL) framework introduces Proxy Fusion to retain privileged knowledge, enabling iterative self-refinement during inference by using intermediate restoration outputs as pseudo-privileged information. Retrained PromptIR with the proposed SIPL achieves significant improvement across diverse all-in-one tasks.}
    \label{fig:learning_frameworks}
    \vspace{-15pt}
\end{figure}

To address this fundamental limitation, we draw inspiration from Privilege Learning (PL)~\cite{vapnik2009new, vapnik2015learning}. We propose that privileged information, specifically GT-derived supervisory signals, can mitigate these adverse effects. The incorporation of privileged information during training establishes an \emph{inter-task comprehension bridge}, effectively harmonizing conflicting gradients and stabilizing convergence. For instance, when we validated a PL-informed training framework on the PromptIR model, it not only achieved performance comparable to the original baseline in less than half the training epochs but also ultimately demonstrated significantly superior results. As illustrated in Figure~\ref{fig:learning_frameworks}(a), this form of privileged guidance demonstrably stabilizes the multi-task learning process, accelerates overall convergence, and enhances generalization across diverse degradation types. Introducing PL therefore presents an efficient and straightforward strategy for improving existing all-in-one restoration methods. Nevertheless, vanilla PL traditionally limits its own impact, as the privileged guidance is often progressively diminished during training and is entirely unavailable at inference time.

\vspace{-5pt}
    
    \vspace{5pt}
In this paper, we identify a critical, often overlooked opportunity: retaining the essence of privileged information to enhance inference-time performance. Addressing this key limitation, we propose the novel \emph{Self-Improved Privilege Learning (SIPL)} framework (illustrated in Figure~\ref{fig:learning_frameworks}(b)). SIPL significantly extends traditional PL paradigms by enabling the model to leverage its intermediate restoration outputs as \emph{pseudo-privileged information}, thereby introducing an iterative self-enhancement mechanism that actively refines performance during inference.
Central to our SIPL approach is the effective \emph{Proxy Fusion} mechanism, which employs a set of learnable parameters termed the \emph{Privileged Dictionary}. Unlike vanilla PL strategies that directly blend privileged and primary features, Proxy Fusion exploits cross-attention to distill high-quality image priors from GT-derived privileged features during training. Critically, the learned Privileged Dictionary persists beyond the training phase. At inference, initial restoration outputs, although imperfect, inherently contain structural information significantly closer to the GT than the original degraded inputs. These outputs subsequently serve as pseudo-privileged information, iteratively interacting with this pre-learned Privileged Dictionary. This iterative refinement progressively improves internal feature representations, facilitating continuous, self-driven enhancement. The efficacy of this approach is evident from our validations: when integrated with the PromptIR model, SIPL yields substantial improvements across multiple all-in-one image restoration benchmarks, achieving PSNR gains of +4.58~dB on Composite Degradation, +1.23~dB on Allweather, and +1.38~dB on the Five-Task settings.
    
    \vspace{-2pt}
Our primary contributions are summarized as follows:
\begin{itemize}
    \vspace{-5pt}

    \item We analyze the optimization impediments in all-in-one image restoration and establish the efficacy of Privilege Learning as a potent foundational strategy, demonstrating its ability to mitigate task competition and improve multi-task convergence.
    \vspace{-5pt}
    
    \item We propose the novel Self-Improved Privilege Learning (SIPL) framework, which uniquely extends PL to enable iterative self-improvement during inference.
    \vspace{-5pt}
    
    \item We introduce the Proxy Fusion module to efficiently distill and retain knowledge from privileged information. This plug-and-play mechanism adds minimal computational overhead while providing substantial quality improvements when integrated with various backbone restoration architectures.
\end{itemize}

\vspace{-10pt}
\section{Related Work}
\vspace{-10pt}
\subsection{All-in-One Image Restoration}
\label{subsec:all_in_one_ir}
\vspace{-7pt}

Traditional image restoration typically targets specific degradations, such as noise or blur, using specialized models~\cite{zamir2022restormer, huang2023memroyderain, cai2023retinexformer}. However, real-world scenarios often involve unknown or mixed degradations, driving the need for \emph{all-in-one} models that handle diverse degradation types in a unified framework~\cite{su2022survey, jiang2024survey}.

Early efforts toward unified models utilized powerful backbones~\cite{chen2022simple, chen2021IPT, Uformer, zamir2022restormer, wang2024gridformer, guo2024mambair} or generative approaches like diffusion models~\cite{Belhasin2024PUIR, Yue2024Difface, li2024latent, yue2025ResShift}. A key challenge is guiding a single network to handle diverse restoration tasks, leading to a focus on conditioning mechanisms, from early degradation encoders~\cite{Li22AirNet} to advanced learnable prompts~\cite{Potlapalli2023promptir}, sometimes refined with frequency priors or dimensionality reduction~\cite{cui2025adair, zhang2023IDR}. Other methods use degradation priors from classifiers~\cite{MioIR} or vision-language models like CLIP with textual prompts~\cite{Luo2024DA-CLIP, Lin2024Textual}. Additionally, multi-task learning techniques~\cite{wu2024harmony, wu2025debiased} and pretraining strategies~\cite{qin2024maskedIR} address optimization challenges. While these advances improve all-in-one capabilities, they often focus on architectural or input conditioning changes, overlooking optimization stability in multi-task learning.

\vspace{-7pt}
\subsection{Privilege Learning}
\label{subsec:privilege_learning}\vspace{-8pt}
Privilege Learning (PL), formally introduced as Learning Using Privileged Information (LUPI)~\cite{vapnik2009new, vapnik2015learning}, is a machine learning paradigm where auxiliary information, inaccessible during inference, is provided to the model during training. This privileged information acts as a form of expert guidance, aiming to simplify the learning task, improve convergence, or enhance generalization, even though the model must perform inference without it. The core concept is that this richer training signal can help the model learn a more robust internal representation or decision boundary.

\vspace{-10pt}
\section{Method}
\vspace{-10pt}
\label{sec:method}
In this section, we present our Self-Improved Privilege Learning (SIPL) framework for all-in-one image restoration. We first establish the foundational Privilege Learning (PL) paradigm to image restoration. We then introduce our novel SIPL, which uniquely enables self-improvement during inference through our proposed Proxy Fusion mechanism.

\vspace{-7pt}
\subsection{Privilege Learning}\label{method:privilege_learning}
\vspace{-7pt}

As established in Section 1, all-in-one image restoration faces fundamental optimization challenges stemming from task competition and conflicting gradient directions. Privilege Learning~\cite{vapnik2009new, vapnik2015learning} offers an elegant solution by leveraging additional information during training that enhances the learning process. Given a degraded image $I_d$ and its corresponding ground truth $I_{gt}$, PL incorporates privileged information (derived from $I_{gt}$) into the training process. This is achieved through a simple yet effective feature fusion approach:

\vspace{-10pt}
\begin{equation}
F_{fused} = (1-\alpha) \cdot F_d + \alpha \cdot F_{PI},
\label{eq:pl_fusion}
\end{equation}
\vspace{-10pt}

where $F_d$ represents features extracted from the degraded image, $F_{PI}$ denotes features derived from the ground truth (privileged information), and $\alpha \in [0,1]$ controls the degree of privileged guidance. During training, $\alpha$ typically follows a decreasing schedule, ensuring the model gradually adapts to operating without privileged guidance. At inference time, $\alpha$ is set to 0, as privileged information is unneeded.

This straightforward mechanism provides substantial benefits for all-in-one restoration. It stabilizes training and mitigates task competition by providing clear guidance, particularly for challenging degradation types. Meanwhile, our introduced PL paradigm remains agnostic to model architecture, enhancing all-in-one image restoration generally.

\vspace{-7pt}
\subsection{Self-Improved Privilege Learning}
\label{subsec:sipl}
\vspace{-7pt}

Beyond the vanilla PL paradigm, we propose a novel extension to retain the privileged information at the inference stage. Our key insight is that the restoration output, though imperfect, contains valuable information closer to the ground truth than the original degraded input. This observation motivates our proxy fusion module, which enables the model to leverage its own outputs as pseudo-privileged information during inference.

\vspace{-0.55cm}
\begin{wrapfigure}[22]{r}{0.3\textwidth}
 \centering
\includegraphics[width=\linewidth]{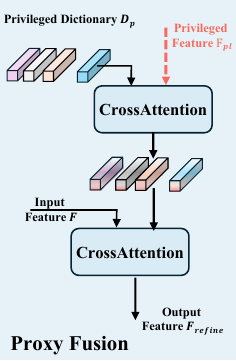}  
\caption{Implementation of the proxy fusion module.}
\label{fig:framework_overview}  
\end{wrapfigure}

\paragraph{Proxy Fusion}
The cornerstone of SIPL is our novel Proxy Fusion mechanism, which creates a persistent bridge between training-time privileged knowledge and inference-time self-improvement. Unlike the direct feature blending in conventional PL (Eq.~\ref{eq:pl_fusion}), Proxy Fusion employs a learnable Privileged Dictionary (PD) to distill and retain essential knowledge from privileged information:

\vspace{-5pt}

\begin{equation}
\text{PD} \in \mathbb{R}^{N \times C},
\end{equation}
\vspace{-5pt}

where $N$ represents the number of dictionary entries and $C$ is the feature dimension. This dictionary interacts with privileged features through cross-attention:

\begin{equation}
F'_{PI} = \text{CrossAttention}(Q=\text{PD}, K=F_{PI}, V=F_{PI}),
\end{equation}

This interaction allows the PD to capture and internalize high-frequency details and statistical patterns characteristic of high-quality images. The distilled knowledge is then integrated with features from the degraded image:

\begin{equation}
F_{fused} = \text{CrossAttention}(Q=F_d, K=F'_{PI}, V=F'_{PI}).
\end{equation}

During training, the entire framework, including the Privileged Dictionary, learns to extract and utilize privileged information effectively. Crucially, the learned PD persists into inference, serving as a knowledge repository that enables self-improvement without requiring actual ground truth.

\paragraph{Self-Improvement Mechanism}
The distinguishing feature of SIPL is its iterative self-refinement capability during inference:

\begin{enumerate}
    \item \textbf{Initial Restoration}: The model produces an initial output $I_{restored}^{(0)}$ using only the degraded input:
    \begin{equation}
    I_{restored}^{(0)} = \mathcal{F}(I_d).
    \end{equation}
    
    \item \textbf{Self-Improvement}: This initial output serves as pseudo-privileged information. Features extracted from $I_{restored}^{(0)}$ interact with the learned PD through the Proxy Fusion mechanism, guiding subsequent restoration:
    \begin{equation}
    I_{restored}^{(1)} = \mathcal{F}(I_d, I_{restored}^{(0)}).
    \end{equation}
    
    \item \textbf{Iterative Refinement (Optional)}: This process can continue for multiple iterations, with each step potentially improving quality:
    \begin{equation}
    I_{restored}^{(t)} = \mathcal{F}(I_d, I_{restored}^{(t-1)}), \quad t \geq 1.
    \end{equation}
\end{enumerate}

This elegant self-correction loop enables progressive quality enhancement without requiring actual ground truth during deployment. In practice, we find that a single refinement step ($t=1$) typically provides substantial improvements with minimal computational overhead.

The key advantage of our Proxy Fusion approach over direct feature blending is its ability to distill and retain essential high-quality image characteristics in the learned PD parameters. This creates a persistent knowledge repository that facilitates self-improvement during inference, a capability absent in conventional PL frameworks.

\vspace{-5pt}
\subsection{Remarks}
\label{subsec:remarks}
\vspace{-7pt}

Our SIPL framework focuses exclusively on the learning paradigm rather than specific architectural choices. The Proxy Fusion module serves as a plug-and-play component that can enhance virtually any existing restoration architecture. This architectural agnosticism ensures broad applicability across diverse restoration models and tasks. For experimental validation, we integrate SIPL into multiple distinct backbone architectures, including PromptIR~\cite{Potlapalli2023promptir}. As demonstrated in Section~\ref{sec:experiments}, SIPL consistently improves performance across all tested models, confirming its generality and effectiveness as an optimization framework for all-in-one image restoration.

\vspace{-7pt}

\section{Experiments}\label{sec:experiments}
\vspace{-10pt}

In this section, we conduct extensive experiments to evaluate our Self-Improved Privilege Learning (SIPL) framework across various all-in-one image restoration settings. We first describe our experimental setup, implementation details, and evaluation metrics. Then, we present quantitative and qualitative comparisons with state-of-the-art methods, followed by comprehensive ablation studies that validate the effectiveness of our key components.

\begin{table*}[!t]
\setlength{\abovecaptionskip}{0pt}
\caption{Quantitative results (PSNR/SSIM) on the Three-Task Setting. Our results are highlighted in \textbf{bold}, and best results are \underline{underlined}.}
\label{tab:three_task}
  \centering
\large
\renewcommand\arraystretch{1.15}
    \resizebox{\linewidth}{!}{\begin{tabular}{c|l|l|ccc|c|c|c} % \resizebox{\linewidth}{!}{
    \toprule[1pt]
 \multirow{2}{*}{\bf Type}& \multirow{2}{*}{\bf Method} &  \multirow{2}{*}{\bf Venue} 
    & \multicolumn{3}{c|}{\textbf{Denoising} (BSD68)}
    & \multicolumn{1}{c|}{\bf Dehazing}
    & \multicolumn{1}{c|}{\bf Deraining}
    & \multirow{2}{*}{\bf Average}
 
    \\ \cline{4-8}  % \cmidrule(r){3-5} \cmidrule(r){6-6} \cmidrule(r){7-7}

   & &  & $\sigma = 15$ & $\sigma = 25$ & $\sigma = 50$  
    & SOTS 
    & Rain100L  
    &  \\
    \hline
  \multirow{6}{*}{\rotatebox{90}{\textit{General}}} & MPRNet \cite{Zamir_2021_CVPR_mprnet} & CVPR'21
    & 33.27/0.920  & 30.76/0.871  & 27.29/0.761    & 28.00/0.958   & 33.86/0.958  & 30.63/0.894 \\
   & Restormer \cite{zamir2022restormer} & CVPR'22 
    & 33.72/{0.930}  & 30.67/0.865  & 27.63/{0.792}    & 27.78/0.958   & 33.78/0.958  & 30.75/0.901 \\
  & NAFNet \cite{chen2022simple} & ECCV'22 
    & 33.03/0.918  & 30.47/0.865  & 27.12/0.754    & 24.11/0.928   & 33.64/0.956  & 29.67/0.844  \\
   & FSNet* \cite{cui2022SFNet}  &  TPAMI'23 
    & 33.81/0.930   &30.84/0.872   & 27.69/0.792     & 29.14/0.968    & 35.61/0.969   & 31.42/0.906   \\
    & DRSformer* \cite{chen2023drsformer}     &    CVPR'23 
    &33.28/0.921   &30.55/0.862   &27.58/0.786     &29.02/0.968    & 35.89/0.970   &{31.26}/0.902 \\
   & MambaIR* \cite{guo2024mambair} & ECCV'24
    & 33.88/0.931  &  30.95/0.874  & 27.74/0.793    &29.57/0.970   & 35.42/0.969 & 31.51/0.907 \\  
    \hline
  \multirow{9}{*}{\rotatebox{90}{\textit{All-in-One}}} &  DL \cite{dl} & TPAMI'19
    & 33.05/0.914  & 30.41/0.861  & 26.90/0.740    & 26.92/0.391   & 32.62/0.931  & 29.98/0.875 \\
   & AirNet \cite{Li22AirNet} & CVPR'22 
    & 33.92/{0.932}  & 31.26/{0.888}  & 28.00/0.797    & 27.94/0.962   & 34.90/0.967  
    & 31.20/0.910  \\
    & IDR* \cite{zhang2023IDR}  & CVPR'23 
    & {33.89}/0.931  & {31.32}/0.884  & {28.04}/0.798    & 29.87/0.970   & 36.03/0.971  & 31.83/0.911  \\

  & Gridformer* \cite{wang2024gridformer}  &  IJCV'24
    & 33.93/0.931  & 31.37/0.887  & 28.11/0.801    
    & 30.37/0.970   & 37.15/0.972  
    & 32.19/0.912  \\
  & NDR \cite{yao2024ndr}  & TIP'24 
    & {34.01}/0.932  & {31.36}/0.887  & {28.10}/0.798    & 28.64/0.962   
    & 35.42/0.969  & 31.51/0.910 \\
 & InstructIR \cite{Conde2024InstructIR}  & ECCV'24 
    & \underline{34.15}/{0.933}  &\underline{31.52}/{0.890}  & \underline{28.30}/\underline{0.804} 
    & {30.22}/{0.959}   & {37.98}/{0.978}  
    & {32.43}/{0.913} \\
  & TextualDegRemoval\cite{Lin2024Textual} &   CVPR'24
    & {34.01}/0.933 & {31.39/0.890}  
    & {28.18}/{0.802}  & \underline{31.63}/\underline{0.980}   
    & {37.58/0.979}  & {32.63/0.917} \\
& AdaIR \cite{cui2025adair}
& ICLR'25
& 34.12/\underline{0.935} 
& 31.45/\underline{0.892} 
& 28.19/0.802 
& 31.06/0.980 
& \underline{38.64}/0.983
& \underline{32.69}/\underline{0.918}
  \\
\cline{2-9}
  & PromptIR \cite{Potlapalli2023promptir}   
  & NeurIPS'23
    & 33.98/{0.933}  
    & 31.31/{0.888} 
    & 28.06/{0.799}    
    & {30.58}/{0.974}   
    & {36.37}/{0.972}  
    & {32.06}/{0.913}  \\
&\cellcolor{my_color}\textbf{PromptIR + SIPL}  
&\cellcolor{my_color} 2025
& \cellcolor{my_color}\textbf{34.12}/\textbf{0.933}
& \cellcolor{my_color}\textbf{31.48}/\textbf{0.889} 
& \cellcolor{my_color}\textbf{28.22}/\textbf{0.800}
& \cellcolor{my_color}\textbf{31.09}/\textbf{0.977} 
& \cellcolor{my_color}\textbf{38.43}/\underline{\textbf{0.984}}
& \cellcolor{my_color}\textbf{32.67}/\textbf{0.917} \\
  \bottomrule[1pt]
  \end{tabular}
  }
  \vspace{-15pt}
\end{table*}

\vspace{-7pt}

\subsection{Experimental Setup}
\label{subsec:exp_setup}
\vspace{-10pt}

We evaluate SIPL across five comprehensive all-in-one settings that encompass a wide range of real-world image degradation scenarios:

\begin{enumerate} 
\vspace{-7pt}
\item \textbf{Three-Task Setting:} Following the established protocol in \cite{Li22AirNet,Potlapalli2023promptir}, we address three distinct degradation tasks: image denoising (with synthetic Gaussian noise), deraining (using the Rain100L dataset), and dehazing (on the RESIDE dataset). For denoising, we use BSD400 \cite{BSD500} and WED \cite{WED_dataset} for training, and test on BSD68 with noise levels of 15, 25, and 50. For deraining, we employ the Rain100L \cite{rain100} dataset, and for dehazing, we use the outdoor subset of RESIDE \cite{RESIDE_dataset}.

\item \textbf{Five-Task Setting:} To evaluate our model's capacity to handle a broader spectrum of degradations, we utilize a five-task benchmark, including deraining (Rain100L \cite{rain100}), dehazing (RESIDE Indoor Training Set), denoising (BSD400 + WED), motion deblurring (GoPro \cite{gopro_dataset}), and low-light enhancement (LOL \cite{LOL_v1}). This setting follows \cite{zhang2023IDR} and tests the model's ability to handle diverse degradation types in real-world scenarios.

\item \textbf{Deweathering Setting:} Based on previous work \cite{Transweather}, we use the AllWeather \cite{Transweather} dataset for training, containing images from Raindrop \cite{raindrop}, Outdoor-Rain \cite{test1_dataset_allweather}, and Snow100K \cite{Snow100K_dataset}.

\item \textbf{Composite Degradation Setting:} Following the protocol established in \cite{guo2024onerestore}, we evaluate our method on the Composite Degradation Dataset (CDD-11), which represents a more challenging scenario with mixed degradations. CDD-11 encompasses 11 categories of image degradations including single degradations (low-light, haze, rain, snow) and their combinations (low+haze, low+rain, low+snow, haze+rain, haze+snow, low+haze+rain, low+haze+snow). The dataset is constructed using standard benchmarks: the LOw-Light dataset (LOL) \cite{LOL_v1}, the REalistic Single Image DEhazing Outdoor Training Set (RESIDE-OTS) \cite{RESIDE_dataset}, the Rain1200 dataset \cite{rain1200}, and the Snow100k dataset \cite{Snow100K_dataset}. This setting particularly evaluates our framework's capability to handle complex, interacting degradations that better reflect real-world scenarios.
\end{enumerate}
\vspace{-7pt}

For all settings, we adopt the same training/testing splits and protocols as in the original works to ensure fair comparisons. We integrate our proposed SIPL framework into various backbone architectures to demonstrate its versatility and effectiveness.

\begin{table*}[t]
\setlength{\abovecaptionskip}{0pt}
\caption{Quantitative results (PSNR/SSIM) on the Five-Task Setting. Our results are highlighted in \textbf{bold}, and best results are \underline{underlined}.}
\label{tab:five_task}
  \centering
  \large
  \renewcommand\arraystretch{1.15}
    \resizebox{\linewidth}{!}{\begin{tabular}{c|l|l|c|c|c|c|c|c} % \resizebox{\linewidth}{!}{
  
    \toprule[1pt]
 \multirow{2}{*}{\bf Type} & \multirow{2}{*}{\bf Method} & \multirow{2}{*}{\bf Venue} 
    & \multicolumn{1}{c|}{\bf Denoising}
    & \multicolumn{1}{c|}{\bf Dehazing}
    & \multicolumn{1}{c|}{\bf Deraining}
    & \multicolumn{1}{c|}{\bf Deblurring}
    & \multicolumn{1}{c|}{\bf Low-light}
    & \multirow{2}{*}{\bf Average}
    \\ \cline{4-8} 
 &  & & BSD68  & SOTS   & Rain100L   & GoPro   & LOL  &
   \\
    \hline
   \multirow{8}{*}{\rotatebox{90}{\textit{General}}}  &  SwinIR \cite{SwinIR}&  ICCVW'21 
    & 30.59/0.868  & 21.50/0.891  & 30.78/0.923    & 24.52/0.773   & 17.81/0.723  & 25.04/0.835 
  % \\
  %    &MIRNet-v2 \cite{MIRNet_v2}& TPAMI'22  
  %   & 30.97/0.881  & 24.03/0.927  & 33.89/0.954    & 26.30/0.799   & 21.52/0.815  & 27.34/0.875  
 \\
  &Restormer \cite{zamir2022restormer} & CVPR'22 
    & {31.49}/0.884  & 24.09/0.927  & 34.81/0.962    & 27.22/0.829   & 20.41/0.806  & 27.60/0.881  \\

  &NAFNet \cite{chen2022simple} &  ECCV'22 
    & 31.02/0.883  & 25.23/0.939  & 35.56/0.967    & 26.53/0.808   & 20.49/0.809  & 27.76/0.881  \\
  & DRSformer* \cite{chen2023drsformer}&  CVPR'23 
    &30.97/0.881   &24.66/0.931   &33.45/0.953     &25.56/0.780    &21.77/0.821   & 27.28/0.873
 \\
 & Retinexformer* \cite{cai2023retinexformer}  & ICCV'23  
    & 30.84/0.880  &24.81/0.933   &32.68/0.940     &25.09/0.779    &  22.76/0.834   & 27.24/0.873
    \\
  &FSNet* \cite{cui2024FSNet}&  TPAMI'23  
    & 31.33/0.883  &  25.53/0.943  & 36.07/0.968    & 28.32/0.869   & 22.29/0.829  &  28.71/0.898 \\
  &MambaIR* \cite{guo2024mambair} & ECCV'24 
    & 31.41/0.884  &  25.81/0.944  & 36.55/0.971    & 28.61/0.875   & 22.49/0.832  & 28.97/0.901  \\

    \hline
     
   \multirow{9}{*}{\rotatebox{90}{\textit{All-in-One}}} &   DL \cite{dl} & TPAMI'19 
    & 23.09/0.745  & 20.54/0.826  & 21.96/0.762    & 19.86/0.672   & 19.83/0.712  & 21.05/0.743
  \\

  &  TAPE \cite{TAPE} & ECCV'22 
    & 30.18/0.855  & 22.16/0.861  & 29.67/0.904    & 24.47/0.763   & 18.97/0.621  & 25.09/0.801
 \\

   &  Transweather \cite{Transweather}&  CVPR'22 
    & 29.00/0.841  & 21.32/0.885  & 29.43/0.905    & 25.12/0.757   & 21.21/0.792  & 25.22/0.836
 \\

   & AirNet \cite{Li22AirNet} &   CVPR'22 
    & 30.91/0.882  & 21.04/0.884  & 32.98/0.951    & 24.35/0.781   & 18.18/0.735  & 25.49/0.846
 \\

   &  IDR \cite{zhang2023IDR}&  CVPR'23 
    & \underline{31.60}/{0.887}  & 25.24/0.943  & 35.63/0.965    & 27.87/0.846   & 21.34/0.826  & 28.34/0.893  \\

  &  Gridformer* \cite{wang2024gridformer} &  IJCV'24 
    & 31.45/0.885  & 26.79/0.951  & 36.61/0.971    
    & 29.22/0.884   & 22.59/0.831  
    & 29.33/0.904 
 \\ 

 & InstructIR \cite{Conde2024InstructIR} &  ECCV'24 
    & {31.40}/0.887  & {27.10}/{0.956}  & {36.84}/{0.973}    
    &  \underline{29.40}/0.886   &  23.00/0.836  
    & {29.55}/{0.907} 
  \\
 & AdaIR\cite{cui2025adair}
 & ICLR'25
 & 31.35/0.889 
 & \underline{30.53}/\underline{0.978}
 & 38.02/0.981 
 & 28.12/0.858 
 & 23.00/0.845 
 & 30.20/0.910
  \\
\cline{2-9}
  & PromptIR* \cite{Potlapalli2023promptir} 
  &  NeurIPS'23 
    & {31.47}/{0.886}  
    & {26.54}/{0.949}  
    & {36.37}/{0.970}    
    & {28.71}/{0.881}   
    & {22.68}/{0.832}  
    & {29.15}/{0.904} \\

  &\cellcolor{my_color}\textbf{PromptIR + SIPL} 
  &\cellcolor{my_color}2025
  &\cellcolor{my_color}\textbf{31.45}/\underline{\textbf{0.888}}  
  &\cellcolor{my_color}\textbf{30.51}/\textbf{0.975}
  &\cellcolor{my_color}\underline{\textbf{38.09}}/\underline{\textbf{0.982}}    
  &\cellcolor{my_color}\textbf{29.35}/\underline{\textbf{0.886}}   
  &\cellcolor{my_color}\underline{\textbf{23.23}}/\underline{\textbf{0.856}} 
  &\cellcolor{my_color}\underline{\textbf{30.53}}/\underline{\textbf{0.917}} \\
					
  \bottomrule[1pt]
  \end{tabular}
}
\end{table*}
\begin{figure}[!h]

\centering
    \includegraphics[width=\linewidth]{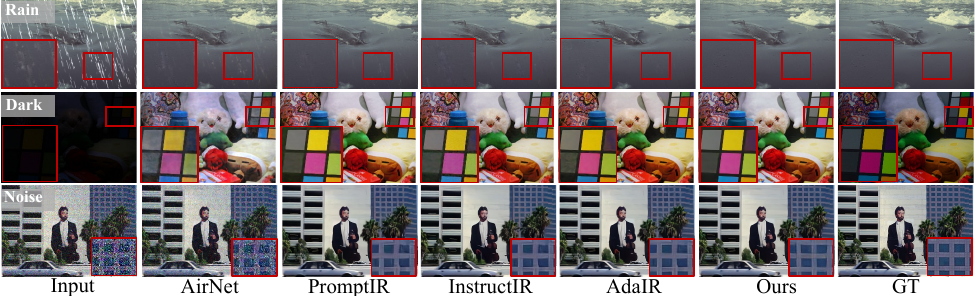}
    \vspace{-15pt}
    \caption{Visual comparison on the Five-Task benchmark. Our method demonstrates superior restoration quality across diverse degradations, effectively recovering finer details and image structures compared to other approaches.}
    \label{fig:five_task}
    \vspace{-5pt}
\end{figure}

\vspace{-10pt}
\subsection{Main Results}
\label{subsec:main_results}
\vspace{-8pt}
We now present the quantitative comparison of our SIPL framework, integrated with the PromptIR backbone (denoted as ``PromptIR + SIPL''), against the original PromptIR and other state-of-the-art methods across the four benchmark settings.

\vspace{-8pt}
\paragraph{Results on Three-Task Setting}
As presented in Table~\ref{tab:three_task}, PromptIR enhanced with our SIPL framework demonstrates marked improvements over the original baseline across all three degradation tasks. Specifically, our method achieves an average PSNR of 32.67 dB, a notable gain of +0.61 dB compared to the PromptIR. The advantages of SIPL are particularly evident in deraining, with a PSNR increase of +2.06 dB (38.43 dB vs. 36.37 dB), and dehazing, with a +0.51 dB PSNR improvement (31.09 dB vs. 30.58 dB). These results position our approach favorably against recent state-of-the-art all-in-one methods, including AdaIR~\cite{cui2025adair} and TextualDegRemoval~\cite{Lin2024Textual}, establishing its strong competitiveness.

\vspace{-8pt}

\paragraph{Results on Five-Task Setting}
Table~\ref{tab:five_task} details the performance on the more comprehensive five-task benchmark. In this more demanding scenario, SIPL's benefits are even more pronounced. Our retrained model achieves an average PSNR of 30.53 dB, significantly surpassing the PromptIR baseline by +1.38 dB. Substantial gains are observed across multiple tasks, notably in dehazing (+3.97 dB PSNR, 30.51 dB vs. 26.54 dB), deraining (+1.72 dB PSNR, 38.09 dB vs. 36.37 dB), and low-light enhancement (+0.55 dB PSNR, 23.23 dB vs. 22.68 dB). Moreover, our SIPL-augmented model consistently outperforms recent leading methods such as AdaIR~\cite{cui2025adair} and InstructIR~\cite{Conde2024InstructIR}, setting a new state-of-the-art for this challenging multi-task evaluation. Visual comparisons, as illustrated in Figure~\ref{fig:five_task}, further corroborate these quantitative findings. Our SIPL-enhanced model consistently produces sharper images with better-preserved textures and structural details across the diverse range of degradations, effectively handling tasks from denoising to low-light enhancement where other methods may falter.

\begin{table*}[!t]
\setlength{\abovecaptionskip}{5pt}
\setlength{\belowcaptionskip}{0pt}
\small
\renewcommand\arraystretch{1} 
    \centering
\caption{Quantitative results (PSNR/SSIM) on the Deweathering Setting. Our results are highlighted in \textbf{bold}, and best results are \underline{underlined}.}
\label{tab:deweathering}
\large
    
\setlength{\tabcolsep}{4pt}
 \renewcommand\arraystretch{1.15}
    \resizebox{\linewidth}{!}{
\begin{tabular}{lccccccccccc}
    \toprule[1pt]
    \multirow{2}{*}{\textbf{Method}} &  \multirow{2}{*}{\textbf{Venue}}  &  \multicolumn{2}{c}{\textbf{Snow100K-S}} & \multicolumn{2}{c}{\textbf{Snow100K-L}} & \multicolumn{2}{c}{\textbf{Outdoor-Rain}} & \multicolumn{2}{c}{\textbf{RainDrop}} & \multicolumn{2}{c}{\textbf{Average}} \\
    
    \cmidrule(r){3-4} \cmidrule(r){5-6} \cmidrule(r){7-8} \cmidrule(r){9-10} \cmidrule{11-12}
    & & PSNR & SSIM & PSNR & SSIM & PSNR & SSIM & PSNR & SSIM & PSNR & SSIM \\
        \midrule
   % \multirow{8}{*}{\rotatebox{90}{\textit{Task-Specific}}} 
  % &  SPANet~\cite{SPANet_WangY0C0L19}   & CVPR'19    & 29.92 & 0.8260 & 23.70 & 0.7930 & -- & -- & -- & -- & -- & -- \\
  %    &    DesnowNet~\cite{Snow100K_dataset}    & TIP'18  & 32.33 & 0.9500 & 27.17 & 0.8983 & -- & -- & -- & -- & -- & -- \\

  %    & HRGAN~\cite{outdoor}     &   CVPR'19  & -- & -- & -- & -- & 21.56 & 0.8550 & -- & -- & -- & -- \\
  %     &   MPRNet~\cite{Zamir_2021_CVPR_mprnet} &  CVPR'21 &    -- & -- & -- & -- & 28.03 & 0.9192 & -- & -- & -- & -- \\
  %     &  AttentiveGAN~\cite{raindrop}      & CVPR'18 & -- & -- & -- & -- & -- & -- & 31.59 & 0.9170 & -- & -- \\
  %    &    IDT~\cite{xiao2022image} & TIP'22  &    -- & -- & -- & -- & -- & -- & 31.87 & 0.9313 & -- & -- \\
  %     &   NAFNet~\cite{chen2022simple} &   ECCV'22  &  34.79 & 0.9497 & 30.06 & 0.9017 & 29.59 & 0.9027 & -- & -- & -- & -- \\
  %     &  Restormer~\cite{zamir2022restormer}      & CVPR'22 & 36.01 & 0.9579 & 30.36 & 0.9068 & 30.03 & 0.9215 & 32.18 & 0.9408 & -- & -- \\
  %       \midrule
  %   \multirow{12}{*}{\rotatebox{90}{\textit{All-in-One}}} 
 All-in-One~\cite{as2020}   & CVPR'20  & -- & -- & 28.33 & 0.8820 & 24.71 & 0.8980 & 31.12 & 0.9268 & 28.05 & 0.9023 \\
 Transweather~\cite{Transweather}    & CVPR'22   & 32.51 & 0.9341 & 29.31 & 0.8879 & 28.83 & 0.9000 & 30.17 & 0.9157 & 30.20 & 0.9094 \\
      % &  Chen~\etal~\cite{ChenHTYDK22} &  CVPR'22  &    34.42 & 0.9469 & 30.22 & 0.9071 & 29.27 & 0.9147 & 31.81 & 0.9309 & 31.43 & 0.9249 \\
  WGWSNet~\cite{weather_data}    &  CVPR'22  & 34.31 & 0.9460 & 30.16 & 0.9007 & 29.32 & 0.9207 & 32.38 & 0.9378 & 31.54 & 0.9263 \\
  WeatherDiff\(_{64}\)~\cite{ozan2023weatherdiff}   & TPAMI'23 & 35.83 & 0.9566 & 30.09 & 0.9041 & 29.64 & 0.9312 & 30.71 & 0.9312 & 31.57 & 0.9308 \\
 WeatherDiff\(_{128}\)~\cite{ozan2023weatherdiff}     & TPAMI`23 & 35.02 & 0.9516 & 29.58 & 0.8941 & 29.72 & 0.9216 & 29.66 & 0.9225 & 31.00 & 0.9225 \\
  AWRCP~\cite{AWRCP_YeCBSXJYCL23}  & ICCV'23  & 36.92 & 0.9652 & 31.92 & 0.9341 & 31.39 & 0.9329 & 31.93 & 0.9314 & 33.04 & 0.9409 \\
 GridFormer~\cite{wang2024gridformer}    & IJCV'24 & 37.46 & 0.9640 & 31.71 & 0.9231 & 31.87 & 0.9335 & 32.39 & 0.9362 & 33.36 & 0.9392 \\
 MPerceiver~\cite{AiHZW024}  & CVPR'24  & 36.23 & 0.9571 & 31.02 & 0.9164 & 31.25 & 0.9246 & 33.21 & 0.9294 & 32.93 & 0.9319 \\
 DTPM~\cite{DTPM_0001CCXQL024}  & CVPR'24  & 37.01&  0.9663 & 30.92 & 0.9174 &  30.99  &  0.9340 & 32.72 & 0.9440 & 32.91 & 0.9404\\
 Histoformer~\cite{Histoformer_SunRGWC24} & ECCV'24 & {37.41} & {0.9656} & {32.16} & {0.9261} & {32.08} & {0.9389} & \underline{{33.06}} & {0.9441} & {33.68} & {0.9437} \\
    \hline
 PromptIR~\cite{Potlapalli2023promptir} & NeurIPS'23 & {36.88} & {0.9643} 
     & {31.34} & {0.9200} 
     & {30.80} & {0.9229} 
     & {32.20} & {0.9359} 
     & {32.80} & {0.9357} \\
 \cellcolor{my_color}\textbf{PromptIR + SIPL}   & \cellcolor{my_color}2025 
    & \cellcolor{my_color}\underline{\textbf{37.91}} & \cellcolor{my_color}\underline{\textbf{0.9673}} 
    & \cellcolor{my_color}\underline{\textbf{32.34}} & \cellcolor{my_color}\underline{\textbf{0.9291}} 
    & \cellcolor{my_color}\underline{\textbf{32.91}} & \cellcolor{my_color}\underline{\textbf{0.9469}} 
    & \cellcolor{my_color}{\textbf{32.99}} & \cellcolor{my_color}\underline{\textbf{0.9462}} 
    & \cellcolor{my_color}\underline{\textbf{34.03}} & \cellcolor{my_color}\underline{\textbf{0.9473}} \\

\bottomrule[1pt]

\end{tabular} 

}
\end{table*}
\begin{figure}[!h]
\vspace{-5pt}
    \centering
    \includegraphics[width=\linewidth]{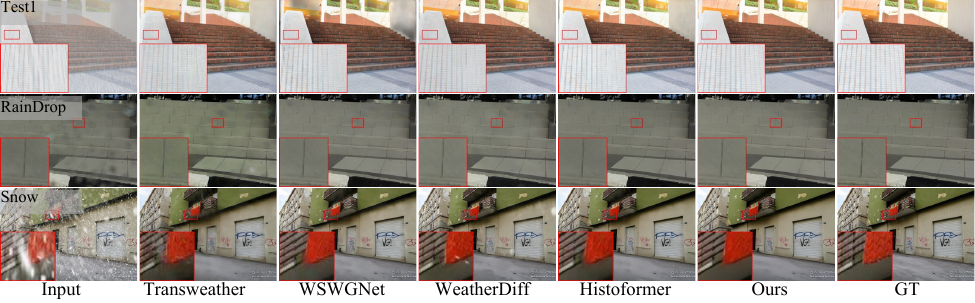}
    \vspace{-17pt}
    \caption{Qualitative examples from the AllWeather dataset. Our method exhibits robust performance in removing various challenging weather conditions. It yields visually superior results with better detail preservation and fewer artifacts.}
    \label{fig:weather}
    \vspace{-7pt}
\end{figure}

\vspace{-10pt}
\paragraph{Results on Deweathering Setting}
The deweathering benchmark, summarized in Table~\ref{tab:deweathering}, evaluates performance on removing diverse adverse weather conditions. Our approach again demonstrates superior capabilities, achieving the new SOTA average PSNR of 34.03 dB and SSIM of 0.9473. This represents a significant +1.23 dB PSNR improvement over the PromptIR baseline. Consistent performance enhancements are recorded across all four test datasets. These results underscore SIPL's robustness in complex deweathering scenarios, outperforming specialized methods and recent deweathering models like Histoformer~\cite{Histoformer_SunRGWC24} and GridFormer~\cite{wang2024gridformer}. The qualitative results presented in Figure~\ref{fig:weather} further highlight the efficacy of our approach. Compared to existing deweathering methods, our retrained PromptIR more effectively removes various weather-related artifacts, such as dense haze and heavy rain, while better preserving image details and avoiding color casts, leading to visually cleaner and more faithful restorations.

\begin{table}[t]
\centering
\caption{Quantitative results (PSNR/SSIM/LPIPS) on the Composite Degradation Setting. Our results are highlighted in \textbf{bold}, and best results are \underline{underlined}.}
\label{tab:composite}
 \renewcommand\arraystretch{1.3}
\vspace{-5pt}
\resizebox{\linewidth}{!}{
\begin{tabular}{cccccccccccccc}
\toprule[1pt]
\textbf{Method} & \textbf{Venue}& \textbf{l} & \textbf{h} & \textbf{r}& \textbf{s}& \textbf{l+h} & \textbf{l+r} & \textbf{l+s}& \textbf{h+r} & \textbf{h+s} & \textbf{l+h+r} & \textbf{l+h+s}  & \textbf{Avg.}\\
\hline
% Restormer &	26.29&	28.35&	33.10&33.43&	24.80&	25.28&	24.99&26.80&	26.15 &23.90 &	23.82 &	26.99 \\
% NAFNet & 24.50 & 25.34 & 29.24 & 29.54 & 21.91 & 22.75 & 22.79 & 23.67 & 23.86 & 21.03 & 20.82 & 24.13 \\
AirNet\cite{Li22AirNet} & CVPR'22 & 24.83 & 24.21 & 26.55 & 26.79 & 23.23 & 22.82 & 23.29 & 22.21 & 23.29 & 21.80 & 22.24 & 23.75  \\
TransWeather\cite{Transweather}&CVPR'22 & 23.39 & 23.95 & 26.69 & 25.74 & 22.24 & 22.62 & 21.80 & 23.10 & 22.34 & 21.55 & 21.01 & 23.13 \\
WeatherDiff\cite{ozan2023weatherdiff} &TPAMI'23 & 23.58 & 21.99 & 24.85 & 24.80 & 21.83 & 22.69 & 22.12 & 21.25 & 21.99 & 21.23 & 21.04  & 22.49 \\
WGWSNet\cite{weather_data}	& CVPR'23& 24.39 &	27.90 &	33.15 & 34.43&	24.27&	25.06&	24.60& 27.23&	27.65&	23.90 & 23.97 &	26.96 \\
InstructIR\cite{Conde2024InstructIR}&ECCV'24 & 26.70 & 32.61 & 33.51 & 34.45 & 24.36 & 25.41 & 25.63 & 28.80 & 29.64 & 24.84 & 24.32 & 28.21 \\

OneRestore\cite{guo2024onerestore}&ECCV'24& 26.55 & 32.71 & 33.48 & 34.50 & 26.15 & 25.83 & 25.56 & 30.27 & 30.46 & 25.18 & 25.28  & 28.47\\
 \hline
PromptIR\cite{Potlapalli2023promptir} & NeurIPS’23 & 26.32 & 26.10 & 31.56 & 31.53 & 24.49 & 25.05 & 24.51 & 24.54 & 23.70 & 23.74 & 23.33 & 25.90\\
% \cellcolor{my_color}\textbf{PromptIR + Pr} 
% & \cellcolor{my_color}2025
% & \cellcolor{my_color}\textbf{27.79} 
% & \cellcolor{my_color}\textbf{30.11} 
% & \cellcolor{my_color}\textbf{33.56} 
% & \cellcolor{my_color}\textbf{34.28} 
% & \cellcolor{my_color}\textbf{26.32} 
% & \cellcolor{my_color}\textbf{25.21}
% & \cellcolor{my_color}\textbf{25.70} 
% & \cellcolor{my_color}\textbf{28.12} 
% & \cellcolor{my_color}\textbf{28.07} 
% & \cellcolor{my_color}\textbf{24.46} 
% & \cellcolor{my_color}\textbf{24.45}
% & \cellcolor{my_color}\textbf{28.01} 
% \\

\cellcolor{my_color}\textbf{PromptIR + SIPL)} 
& \cellcolor{my_color}2025
& \cellcolor{my_color}\underline{\textbf{27.62}} 
& \cellcolor{my_color}\underline{\textbf{36.82}} 
& \cellcolor{my_color}\underline{\textbf{35.66}} 
& \cellcolor{my_color}\underline{\textbf{36.85}} 
& \cellcolor{my_color}\underline{\textbf{27.03}} 
& \cellcolor{my_color}\underline{\textbf{26.79}} 
& \cellcolor{my_color}\underline{\textbf{26.68}}
& \cellcolor{my_color}\underline{\textbf{32.70}} 
& \cellcolor{my_color}\underline{\textbf{32.71}} 
& \cellcolor{my_color}\underline{\textbf{26.20}} 
& \cellcolor{my_color}\underline{\textbf{26.20}} 
& \cellcolor{my_color}\underline{\textbf{30.48}}\\
\bottomrule[1pt]
\hline
\end{tabular}
}
\end{table}
\begin{figure}[!h]
    \centering
    \includegraphics[width=\linewidth]{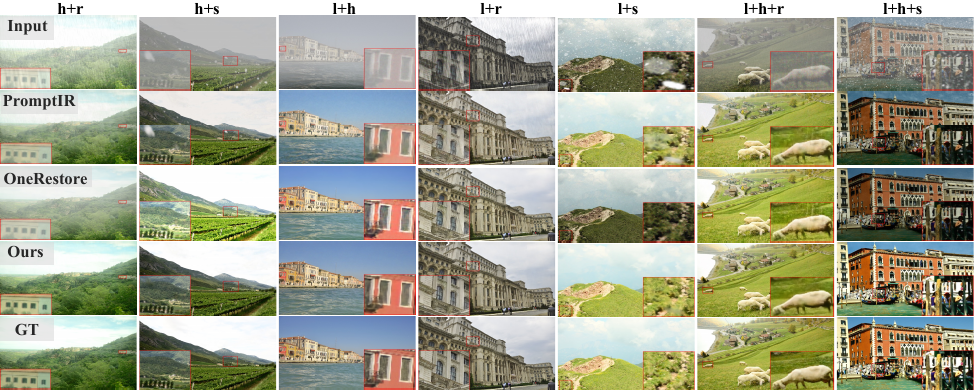}
\caption{Visual results on the composite degradation tasks, showcasing performance on mixed degradations. Our method more effectively mitigates multiple interacting degradations, restoring clearer images with improved color fidelity and detail. }
    \label{fig:cdd}
    \vspace{-15pt}
\end{figure}

\vspace{-10pt}
\paragraph{Results on Composite Degradation Setting}
Table~\ref{tab:composite} evaluates performance on the highly challenging Composite Degradation Dataset (CDD-11), which features images with both single and multiple interacting degradations. Our proposed SIPL delivers outstanding results, achieving an average PSNR of 30.48 dB. This marks a substantial improvement of +4.58 dB over the original PromptIR baseline. This clearly demonstrates the significant contribution of the inference-time self-improvement mechanism, enabled by Proxy Fusion, beyond the benefits of PL-enhanced training alone. The performance leap is particularly remarkable for severe combined degradations: for instance, on haze+snow (h+s) and haze+rain (h+r), our approach achieves PSNR gains of +9.01 dB and +8.16 dB respectively, over the baseline PromptIR. Similarly, for triple degradations such as low-light+haze+snow (l+h+s) and low-light+haze+rain (l+h+r), the improvements are +2.87 dB and +2.46 dB, respectively. Our method also significantly surpasses other recent models designed for composite degradations, such as OneRestore~\cite{guo2024onerestore} and InstructIR~\cite{Conde2024InstructIR}. Figure~\ref{fig:cdd} provides compelling visual evidence of SIPL's advantages on the composite degradation tasks. In scenarios with multiple co-occurring degradations, baseline methods like PromptIR and OneRestore sometimes struggle. For instance with low + snow degradation, they address only one degradation, appearing to remove snow but failing to correct the low-light induced color cast or restore details in dark regions. In contrast, our method demonstrates superior accuracy and generalization.

\begin{wrapfigure}[19]{r}{0.55\textwidth}
\vspace{-0.8cm}
 \centering
 \includegraphics[width=\linewidth]{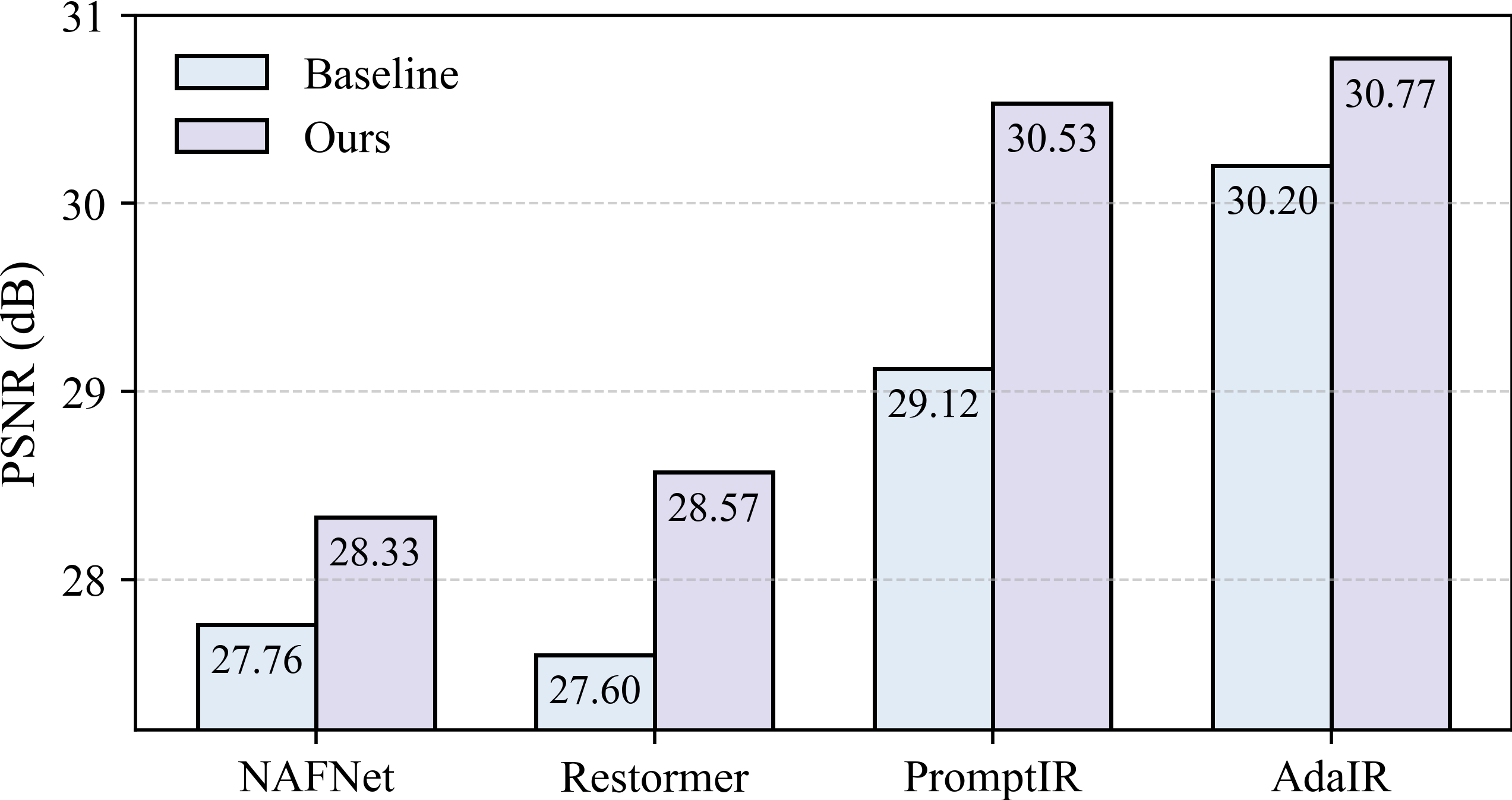} 
    \caption{Ablation study on the interaction of SIPL with diverse backbone architectures on the five-task benchmark. ``Original'' refers to the baseline model, and ``Ours'' refers to the baseline model augmented with our SIPL framework. SIPL consistently improves performance across different architectures.}
    \label{fig:ablation_backbones}
\end{wrapfigure}

\vspace{-10pt}
\subsection{Ablation Studies}
\label{subsec:ablation_studies}
\vspace{-10pt}

In this section, we conduct comprehensive ablation studies to meticulously validate the efficacy of our proposed Self-Improved Privilege Learning framework and dissect the contributions of its core components.

\vspace{-10pt}
\paragraph{Architectural Agnosticism of SIPL}
A core strength of our SIPL framework lies in its architectural agnosticism and ease of integration. To substantiate this plug-and-play capability, we applied SIPL to a spectrum of distinct backbone architectures, moving beyond the PromptIR~\cite{Potlapalli2023promptir}. These included Restormer~\cite{zamir2022restormer}, a prominent Transformer-based network; NAFNet~\cite{chen2022simple}, known for its high CNN efficiency; and AdaIR~\cite{cui2025adair}, a recent state-of-the-art method notable for its frequency domain processing. All models were retrained on the five-task benchmark (as detailed in Section~\ref{subsec:exp_setup}) with and without the SIPL framework integrated. The results, presented in Figure~\ref{fig:ablation_backbones}, unequivocally showcase SIPL's ability to consistently elevate performance across these diverse architectural paradigms. Specifically, integrating SIPL boosted NAFNet's average PSNR from 27.76~dB to 28.33~dB and Restormer's from 27.60~dB to 28.57~dB. The PromptIR model itself saw its performance climb from 29.12~dB to 30.53~dB. Even for a strong baseline like AdaIR, SIPL provided a further enhancement, increasing its average PSNR from 30.20~dB to 30.77~dB. These consistent improvements across varied models underscore SIPL's versatility as a general learning framework, not tightly coupled to any specific architecture, that effectively refines model optimization and enables potent inference-time self-improvement.

\begin{figure}[!t]
    \centering
    \includegraphics[width=\linewidth]{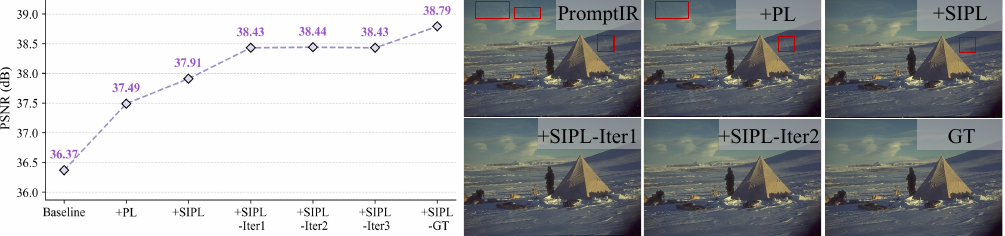}
    \vspace{-10pt}
\caption{Ablation study dissecting the contributions of SIPL's components. The figure illustrates the progressive improvements from the baseline, through the addition of privilege learning (``+PL''), the initial application of SIPL (``+SIPL''), and subsequent iterative self-refinement stages (``+SIPL-IterX''). Performance is benchmarked against an approximate upper bound using GT-guidance (``+SIPL-GT'').}
    \label{fig:ablation_components}
\vspace{-15pt}
\end{figure}

\vspace{-10pt}

\paragraph{Dissecting the Contributions of SIPL Components}
To meticulously evaluate SIPL's core components, we ablate PromptIR on a five degradation tasks, with quantitative and qualitative results presented in Figure~\ref{fig:ablation_components}. The baseline PromptIR model achieves a PSNR of 36.37~dB. Introducing privilege learning solely during training (``+PL'') substantially boosts performance to 37.49~dB. This underscores PL's efficacy in stabilizing multi-task optimization by leveraging privileged information, thereby establishing a stronger foundation. Building upon this, the full SIPL framework, by incorporating the Proxy Fusion module for its initial inference application (``+SIPL''), further elevates the PSNR to 37.91~dB (+0.42~dB over ``+PL''). This increment highlights the Proxy Fusion module's critical role, with its Privileged Dictionary, in effectively distilling, preserving, and transferring privileged knowledge for tangible improvement at inference time. The iterative self-improvement mechanism, a key innovation of SIPL, demonstrates further significant refinement. Crucially, when the model's output from the ``+SIPL'' stage is \textit{first fed back as pseudo-privileged information} (resulting in ``+SIPL-Iter1''), performance impressively surges to 38.43~dB—an additional +0.52~dB gain. This substantial improvement from the initial feedback loop powerfully demonstrates the innovation and effectiveness of using self-generated outputs for refinement. While subsequent iterations show diminishing returns, they affirm the model's capacity for self-correction. This iterative process using pseudo-PI effectively narrows the gap towards the performance upper bound of 38.79~dB, achieved when utilizing true ground truth as privileged input during inference.

\vspace{-10pt}

% \subsection{Limitations and Future Work}
% \label{sec:limitation_future_work}
% \vspace{-10pt}

% Despite SIPL's demonstrated success, further refinements can enhance its utility and address subsisting challenges. A key consideration is the computational cost of iterative refinement. While iterations significantly boost performance, they add inference latency. Although SIPL's initial single-pass application is already potent, future work could focus on optimizing prior distillation and utilization to maximize this initial-pass efficacy. Additionally, the performance gap observed when comparing to GT-guided inference highlights the gap inherent in restored pseudo-privileged information. Enhancing the fidelity of this pseudo-PI or exploiting alignment of privileged dictionary presents a critical research avenue. Finally, a deeper theoretical understanding of SIPL's convergence properties is essential for advancing the overall robustness and broader applicability of such self-improving image restoration approaches.

\subsection{Limitation and Future Work}
\label{sec:limitation_future_work}
\vspace{-10pt}
Our extensive experiments and ablation studies have demonstrated the significant success of the proposed SIPL framework. Nevertheless, we identify key areas and challenges that warrant further investigation. A key consideration is the computational cost of iterative refinement. While iterations significantly boost performance (as shown in Figure~\ref{fig:ablation_components}), they add inference latency. Although SIPL's initial single-pass application is already potent, future work could focus on optimizing prior distillation and utilization to maximize this initial-pass efficacy. Additionally, the performance gap to GT-guided inference underscores the gap of current pseudo-privileged information, which may not fully align with true privileged priors. Thus, improving pseudo-PI fidelity and its alignment with ground truth remains a critical research direction for pushing restoration boundaries. Finally, a more comprehensive theoretical understanding of SIPL's self-improvement convergence is essential for advancing the robustness and broader applicability of all-in-one restoration.

\vspace{-10pt}

\section{Conclusion}
\label{sec:conclusion}
\vspace{-10pt}

In this work, we propose Self-Improved Privilege Learning (SIPL), a novel framework that effectively tackles critical optimization impediments in all-in-one image restoration. SIPL uniquely extends the paradigm of privilege learning to the inference stage: models are empowered to iteratively self-refine their outputs by leveraging them as pseudo-privileged information. This is realized through our proposed proxy fusion module, which employs a privileged dictionary, distilled from ground-truth priors during training, to guide this self-correction process with retrained privileged prior. Extensive evaluations across multiple challenging benchmarks, particularly those with complex composite degradations, confirm that SIPL substantially boosts the performance of diverse state-of-the-art methods, significantly enhancing their robustness and overall restoration quality. We hope that the principles and methodologies presented in SIPL will offer fresh perspectives to the community and stimulate further exploration into more effective strategies for all-in-one image restoration.

\bibliographystyle{unsrtnat}
\bibliography{main}

\newpage
\appendix
\section{Additional Experimental Results and Analyses}
\label{sec:supp_experiments}

This section presents supplementary experimental results, including single-task performance benchmarks, out-of-distribution (OOD) generalization analyses, computational complexity comparisons, further iterative inference studies, and additional qualitative results.

\subsection{Single-Task Performance Evaluation}
\label{sec:supp_single_task}
To further assess the efficacy of \ourmethod{}, we evaluated its performance on individual restoration tasks, aligning our experimental setup with that of \promptir{}~\cite{Potlapalli2023promptir} and \adair{}~\cite{cui2025adair}. These evaluations test the capability of our all-in-one model, enhanced with \ourmethod{}, on specialized degradation scenarios.

\begin{table}[h]
\centering
\caption{Deraining results in the single-task setting on the Rain100L dataset. Our SIPL approach obtains a significant performance boost of 1.98~dB PSNR over baseline PromptIR and 0.12~dB over the AdaIR.}
\label{tab:supp_deraining}
 \renewcommand\arraystretch{1.15}
\resizebox{\linewidth}{!}{
\begin{tabular}{l|cccccccccc}
\toprule[1pt]
 
 Method & DIDMDN & UMR & SIRR & MSPFN & LPNet & AirNet & Restormer & PromptIR & AdaIR & \textbf{PromptIR + SIPL (Ours)} \\
\midrule

PSNR & 23.79 & 32.39 & 32.37 & 33.50 & 33.61 & 34.90 & 36.74 &  {37.04} & {{38.90}} & {\textbf{39.02}} \\
SSIM & 0.773 & 0.921 & 0.926 & 0.948 & 0.958 & 0.977 & 0.978 & {0.979} &  {{0.985}} & {\textbf{0.986}} \\ 
\bottomrule[1pt]
\end{tabular}
}
\end{table}

\paragraph{Deraining on Rain100L:}
The Rain100L dataset~\cite{rain100} serves as a standard benchmark for single-image deraining. As presented in Table~\ref{tab:supp_deraining}, \promptir{} augmented with our \ourmethod{} framework achieves state-of-the-art performance. Specifically, it obtains a PSNR of \textbf{39.02~dB} and an SSIM of \textbf{0.986}. This represents a substantial improvement of \textbf{1.98~dB} in PSNR over the original \promptir{} baseline and also surpasses the strong \adair{} model by \textbf{0.12~dB}, demonstrating the significant benefits of \ourmethod{} in effectively removing rain streaks while preserving image fidelity.

\begin{table}[t]
\centering
\caption{Quantitative comparison for single-task dehazing. Our SIPL achieves significant improvement over baseline PromptIR with 0.51~dB PSNR.}
\label{tab:supp_dehazing}
\resizebox{\linewidth}{!}{%
\begin{tabular}{l|cccccccccc}
    \toprule[1pt]
Method & DehazeNet & MSCNN & AODNet & EPDN &FDGAN & AirNet & Restormer &PromptIR & AdaIR & \textbf{PromptIR + SIPL} \\
 \midrule
    PSNR  &22.46 & 22.06 & 20.29 & 22.57 & 23.15 & 23.18 & 30.87 & {31.31} & {{31.80}} & \textbf{31.82}  \\
    SSIM& 0.851 & 0.908 & 0.877 & 0.863 & 0.921 & 0.900 & 0.969 & {0.973} & {{0.981}} & \textbf{0.982} \\ 
    \bottomrule[1pt]
    \end{tabular}
}
\end{table}

\paragraph{Dehazing on SOTS Outdoor:}
For evaluating dehazing performance, we utilize the outdoor test set from SOTS, part of the RESIDE dataset~\cite{RESIDE_dataset}. The results in Table~\ref{tab:supp_dehazing} indicate that \ourmethod{} notably enhances \promptir{}'s dehazing capabilities. Our approach (\promptir{} + \ourmethod{}) achieves a PSNR of \textbf{31.82~dB} and an SSIM of \textbf{0.982}. This is a gain of \textbf{0.51~dB} in PSNR compared to the \promptir{} baseline. Furthermore, our method slightly outperforms \adair{} (31.80~dB PSNR / 0.981 SSIM), underscoring \ourmethod{}'s efficacy in restoring clarity and detail in hazy conditions.

\begin{table}[h]
\centering
\caption{Image denoising results of directly applying the pre-trained model under the five-degradation setting to the Urban100~\cite{urban100}, Kodak24~\cite{kodak} and BSD68~\cite{bsd68} datasets. The results are PSNR scores. Our SIPL achieves significant improvement across all test datasets compared to previous SOTA method AdaIR\cite{cui2025adair}.}
\label{tab:supp_denoising}
\renewcommand\arraystretch{1.2}

\resizebox{\linewidth}{!}{%
\begin{tabular}{@{}l|ccc|ccc|ccc|c@{}}
\toprule[1pt]
& \multicolumn{3}{c|}{Urban100}& \multicolumn{3}{c|}{Kodak24}  & \multicolumn{3}{c|}{BSD68} \\ 
 Method & $\sigma=15$ &$\sigma=25$&$\sigma=50$ &  $\sigma=15$ &$\sigma=25$&$\sigma=50$ & $\sigma=15$ &$\sigma=25$&$\sigma=50$& Average\\ 
 \hline
DL~\cite{dl}&21.10& 21.28& 20.42& 22.63 &22.66& 21.95& 23.16 &23.09& 22.09  &22.04\\
Transweather~\cite{Transweather}&29.64& 27.97& 26.08& 31.67& 29.64 &26.74&31.16 &29.00 &26.08 &28.66\\
TAPE~\cite{TAPE} &32.19 &29.65& 25.87 &33.24 &30.70& 27.19&32.86 &30.18 &26.63 &29.83\\
AirNet~\cite{Li22AirNet} &33.16 &30.83 &27.45& 34.14 &31.74 &28.59&33.49& 30.91& 27.66 &30.89\\
IDR~\cite{zhang2023IDR} &{33.82}& {31.29} &{28.07}& {34.78}& {32.42}& {29.13}&{34.11}& {31.60} &{28.14}&{31.48}\\ 
AdaIR & {34.10} & {31.68} & {28.29} &{34.89} & {32.38} & {29.21} &  {34.01} & {31.35} & {28.06} &{31.55}\\ 

\textbf{PromptIR + SIPL} & \textbf{35.39} & \textbf{32.74} & \textbf{29.17} &  \textbf{34.98} & \textbf{32.50} & \textbf{29.36} & \textbf{34.08} & \textbf{31.45} & \textbf{28.16} & \textbf{31.98} \\ 

\bottomrule[1pt]
\end{tabular}
}
\end{table}

\paragraph{Denoising using Five-Task Pre-trained Model:}
To assess robustness and generalization for denoising, we employed the all-in-one model pre-trained on five distinct degradation tasks (including denoising) and evaluated it directly on three commonly used denoising benchmark datasets: Urban100~\cite{urban100}, Kodak24~\cite{kodak}, and BSD68~\cite{bsd68}. This setup tests the model's ability to denoise effectively without task-specific fine-tuning. As detailed in Table~\ref{tab:supp_denoising}, \promptir{} + \ourmethod{} demonstrates superior performance, achieving an average PSNR of \textbf{31.98~dB} across all datasets and noise levels ($\sigma \in \{15, 25, 50\}$). This is a notable improvement over \adair{} (31.55~dB average PSNR). Particularly on the Urban100 dataset, which often contains complex structures, our method shows significant gains (e.g., \textbf{+1.29~dB} for $\sigma=15$, \textbf{+1.06~dB} for $\sigma=25$). Consistent, positive improvements are also observed across the Kodak24 and BSD68 datasets for all noise levels. These results, especially on datasets potentially unseen during the denoising phase of the five-task training, highlight the advanced robustness and generalization capabilities endowed by the \ourmethod{} framework.

\paragraph{Summary of Single-Task Evaluations:}
The collective results from these single-task evaluations on deraining, dehazing, and denoising consistently underscore the advantages of integrating \ourmethod{} with a strong baseline like \promptir{}. Across all three distinct restoration tasks, our approach not only yields substantial improvements over the original backbone but also demonstrates competitive or superior performance against recent state-of-the-art methods such as \adair{}. The denoising experiments, in particular, where a versatile five-task model was directly applied to specialized denoising benchmarks, strongly emphasize the generalization power of \ourmethod{}. By enabling more effective utilization of learned priors and facilitating self-improvement, \ourmethod{} significantly enhances the model's robustness and its capability to handle diverse image degradations effectively, even when tested in focused, single-task scenarios.

\begin{table}[h]
\centering
\caption{OOD performance on Rain100L with Gaussian noise ($\sigma=$15, 25, and 50). Models were trained on three distinct tasks and test on this unseen dataset directly. Iter-N denotes N iterative refinement steps. Best results are \underline{underlined}, our method is highlighted.}
\label{tab:supp_ood_rain_noise}
\renewcommand\arraystretch{1.2}
\resizebox{0.7\linewidth}{!}{%
\begin{tabular}{c|ccc|c}
\toprule[1pt]
& \multicolumn{3}{c|}{Rain100L + Gaussian Noise} & \\ 
 Method & $\sigma=15$ &$\sigma=25$&$\sigma=50$ &   Average\\ 
 \hline
PromptIR~\cite{Potlapalli2023promptir} & 24.92 & 24.50 & 23.79 &24.40 \\

AdaIR~\cite{cui2025adair} & 24.91 & 24.50 & 23.77  & 24.39 \\
\hline
\textbf{PromptIR + SIPL (initial)}  & \textbf{24.95} & \textbf{24.59} & \textbf{23.86} & \textbf{24.46} \\
\textbf{+ Iter-1} & \textbf{26.90} & \textbf{25.59} & \textbf{23.96} & \textbf{25.48}  \\
\textbf{+ Iter-2} & \textbf{31.74} & \textbf{29.53} & \textbf{26.41} & \textbf{29.23}
\\

\bottomrule[1pt]
\end{tabular}
}
\end{table}

\begin{figure}[h]
    \centering
    \includegraphics[width=\linewidth]{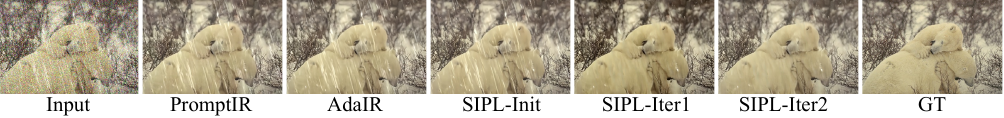}
 \caption{Visual illustration of OOD performance on a challenging Rain100L + Gaussian Noise ($\sigma=50$) example. From left to right: Degraded Input, \promptir{}, \adair{}, \ourmethod{} (Initial), \ourmethod{} (Iter 1), \ourmethod{} (Iter 2), and Ground Truth (GT). The iterative application of \ourmethod{} progressively enhances image clarity, restores fine details, and reduces artifacts, significantly outperforming baseline methods and demonstrating effective generalization to unseen composite degradations.}
 \label{fig:ood_vision}
\end{figure}

\subsection{Out-of-Distribution (OOD) Generalization Analysis}
\label{sec:supp_ood}
A critical attribute of advanced image restoration models is their capacity to generalize effectively to previously unseen degradation types. This section investigates the OOD generalization capabilities of our \ourmethod{}-enhanced framework, drawing comparisons with established methods like \promptir{} and \adair{}. Our analysis specifically focuses on performance when encountering complex, composite degradations not present during the training phase.

\paragraph{Performance on Unseen Composite Degradation (Rain100L + Noise):}
We evaluate models originally trained on three distinct restoration tasks (deraining, dehazing, denoising individually) on a challenging synthetic dataset: Rain100L combined with varying levels of Gaussian noise ($\sigma \in \{15, 25, 50\}$). This composite degradation scenario was deliberately excluded from the training set to rigorously test OOD performance.

The quantitative results are presented in Table~\ref{tab:supp_ood_rain_noise}. Baseline models, \promptir{} and \adair{}, achieve average PSNR scores of 24.40~dB and 24.39~dB, respectively, on this unfamiliar task. Our \promptir{} + \ourmethod{} model, in its initial single-pass inference (``PromptIR + SIPL (initial)''), yields a comparable average PSNR of 24.46~dB. However, the transformative advantage of \ourmethod{} becomes strikingly evident through its iterative self-improvement mechanism. With just one iteration (``+Iter 1''), the average PSNR significantly jumps to 25.48~dB. A second iteration (``+Iter 2'') further elevates the performance dramatically to an average PSNR of \textbf{29.23~dB}. This represents a remarkable \textbf{+4.77~dB} improvement over its initial state and far surpasses the static performance of the baseline models. This progressive and substantial enhancement underscores the robust OOD generalization conferred by \ourmethod{}, particularly its ability to iteratively refine results when faced with novel degradations.

The qualitative improvements are visualized in Figure~\ref{fig:ood_vision} using an example from the Rain100L + Noise ($\sigma=50$) set. While the input image exhibits significant degradation, and baseline methods like \promptir{} and \adair{} offer limited restoration, our \ourmethod{} demonstrates clear visual enhancements. The initial output (``SIPL-Init'') shows some improvement, but subsequent iterations (``SIPL-Iter1'', ``SIPL-Iter2'') progressively recover finer details, enhance sharpness, and reduce artifacts more effectively, approaching the ground truth quality. This visual evidence corroborates the quantitative gains and highlights the practical efficacy of iterative refinement in complex OOD scenarios.

\paragraph{Efficacy of Self-Improvement in OOD Contexts:}
The marked success of \promptir{} + \ourmethod{} in handling these unseen composite degradations, especially through iteration, is attributable to its core design featuring the Privileged Dictionary (PD) and the self-improvement learning strategy. Unlike baseline models such as \promptir{} and \adair{}, which are not inherently designed to leverage their own outputs for iterative refinement on OOD tasks without a guiding mechanism, \ourmethod{} excels in this regard. Standard architectures, if naively iterated on OOD inputs, might see performance stagnate or even degrade due to the accumulation of errors or model biases when processing unfamiliar data distributions.

In stark contrast, \ourmethod{}'s PD, trained to distill essential characteristics of high-quality images, provides robust guidance even when the pseudo-privileged information is derived from an imperfectly restored OOD image. The iterative process allows the model to progressively correct errors and enhance image quality by repeatedly consulting these learned priors. This capacity for effective self-correction and refinement in the face of novel, complex degradations is a key differentiator of our approach.

This OOD analysis strongly suggests that our self-improved iteration paradigm offers a novel and potent pathway for advancing all-in-one image restoration. Beyond striving for optimal performance in a single forward pass, \ourmethod{} demonstrates the significant potential of empowering models to adapt and improve their outputs dynamically at test time. This is particularly crucial for real-world scenarios where diverse and unforeseen degradations are common, showcasing a promising direction for developing more versatile and robust restoration solutions.

\begin{table}[h]
\centering
\caption{Comparison of model parameters and computational complexity.}
\label{tab:supp_complexity}
\resizebox{0.8\linewidth}{!}{%
\begin{tabular}{l c c c }
\toprule
Method & Parameters  & FLOPs & Avg. (dB) \\
\midrule
AirNet~\cite{Li22AirNet} & 9M & 301G & 25.44\\ % Placeholder
Transweather~\cite{Transweather} & 21.5 & 115.2 & 25.22 \\ % Placeholder
\promptir{}~\cite{Potlapalli2023promptir} & 36M & 173G  & 29.15 \\ % Placeholder
\adair{}~\cite{cui2025adair} & 29M & 162G  & 30.20 \\ % Placeholder
\midrule

 \textbf{\promptir{} + PL} & \textbf{36M} & \textbf{173G} & \textbf{30.05} \\ 
 \textbf{\promptir{} + \ourmethod{}} (Iter-0 / Single Pass) & \textbf{39M} & \textbf{193G} & \textbf{30.17} \\ 
 \textbf{\promptir{} + \ourmethod{}} (Iter-1 / Ours) & \textbf{39M} & \textbf{434G} & \textbf{30.53} \\  
\bottomrule
\end{tabular}%
}
\end{table}
\subsection{Efficiency Evaluation}
\label{sec:supp_complexity}
We evaluate the parameter count and computational complexity (FLOPs, calculated for a $256 \times 256$ input) of our proposed methods against several baselines, alongside their average PSNR performance (dB) on the five-task benchmark. The results are summarized in Table~\ref{tab:supp_complexity}.

Our investigation begins with the integration of Privilege Learning (PL) into \promptir{} (denoted as ``\promptir{} + PL''). As a training strategy, PL enhances the model's learning for multi-degradation scenarios. Notably, this approach achieves an average PSNR of 30.05~dB. Comparing to the baseline \promptir{} (29.15~dB), this represents an improvement of 0.9~dB. Critically, this performance boost is achieved while maintaining the original model's parameter count (36M) and FLOPs (173G) during inference, as PL primarily modifies the training process. This underscores PL's efficacy as a zero-cost inference enhancement for all-in-one image restoration.

Building upon this, our Self-Improved Privilege Learning (\ourmethod{}) framework, even in a single pass (``\promptir{} + \ourmethod{} (Iter-0)''), yields further performance gains, reaching 30.17~dB. This increment validates the effectiveness of our proposed Privileged Dictionary (PD) within the Proxy Fusion module. The PD successfully captures essential structural information during training and, crucially, retains this privileged knowledge for inference, thereby extending the conventional PL paradigm. The introduction of the Proxy Fusion module and the privileged feature handling results in a marginal increase of 3M parameters and 20G FLOPs over the ``\promptir{} + PL'' model. This is because the acquisition of privileged features for Proxy Fusion is efficient, primarily leveraging encoder-level features rather than requiring full model duplication for the privileged path.

The true potential of \ourmethod{} is unlocked through iterative inference (``\promptir{} + \ourmethod{} (Iter-1)''), which achieves a higher PSNR of 30.53~dB, demonstrating self-driven performance enhancement. We acknowledge that this iterative process leads to a considerable increase in computational load (434G FLOPs) compared to single-pass models. However, this increased cost is often justified by the substantial and consistent performance improvements observed across diverse restoration tasks and, significantly, in challenging Out-of-Distribution (OOD) scenarios as detailed in Section~\ref{sec:supp_ood}. The self-improved iteration capability provides a powerful form of test-time adaptation, offering a highly effective solution for handling unknown or complex composite degradations in the all-in-one image restoration landscape.

Looking ahead, while the current iterative framework demonstrates compelling results, future research will focus on optimizing its efficiency. We plan to explore avenues such as simplifying the overall design, potentially by identifying and leveraging key model weights more selectively. Further investigation into using the Privileged Dictionary more sparsely within specific feature spaces could also reduce the computational demands of the privileged feature fusion process, aiming to strike an even better balance between performance and efficiency.

\end{document}